\documentclass[lettersize,journal]{IEEEtran}
\usepackage{amsmath,amssymb,amsfonts,mathtools,bm}
\usepackage{graphicx}
\usepackage{cite}
\usepackage{booktabs}
\usepackage{array}
\usepackage{pgfplots}
\usepackage{algorithm}
\usepackage{algorithmic}
\usepackage{hyperref}
\hypersetup{hidelinks}
\pgfplotsset{compat=1.18}
\usepgfplotslibrary{groupplots}
\usepackage{xcolor}
\graphicspath{{figure/}}
\usepackage{multirow}

\newcommand{\R}{\mathbb{R}}
\newcommand{\SE}{SE(3)}

\newcommand{\I}{\bm{I}}
\newcommand{\zero}{\bm{0}}
\newcommand{\diag}{\mathrm{diag}}
\newcommand{\Ad}{\mathrm{Ad}}
\newcommand{\adop}{\mathrm{ad}}

\begin{document}
	
	\title{Design, Modeling and Experimental Validation of a Miniature Hybrid Underwater Glider\\
		With Large-Range Foldable Deflectable Wings}
	
	\author{Yongjian Zhu, Yusen Tao and Feitian Zhang%
        \thanks{The authors are with the Robotics and Control Laboratory, School of Advanced Manufacturing and Robotics, and the State Key Laboratory of Turbulence and Complex Systems, Peking University, Beijing 100871, China (e-mail: yongjianzhu@pku.edu.cn; taoyusen@stu.pku.edu.cn; feitian@pku.edu.cn).}}
	\maketitle
	
	\begin{abstract}		
		Miniature hybrid underwater gliders have attracted increasing attention for long-endurance ocean observation and confined-space inspection. Large-range wing reconfiguration offers a promising yet largely unexplored approach for simultaneously enhancing maneuverability and shape adaptability in constrained underwater environments. However, such morphing introduces substantial challenges in mechanical integration, dynamic modeling, and hydrodynamic characterization. This paper presents FoDeGlider, a miniature hybrid underwater glider equipped with two independently actuated wings capable of large-range folding and deflection. To  capture configuration-dependent variations in mass distribution, center-of-geometry location, and hydrodynamic loading, a multibody dynamics model is developed by treating wing configuration as a structural variable. A composite rigid body algorithm (CRBA)-based projection formulates the composite inertia, wrench transformations, and component-level hydrodynamics into a unified Fossen-form dynamic model applicable to arbitrary wing configurations. A sequential parameter-identification framework is further proposed to estimate fuselage and wing hydrodynamic coefficients, resulting in an open benchmark dataset for model identification and validation. Extensive experiments are conducted, the results of which demonstrate accurate dynamic modeling and parameter identification across diverse morphing configurations. Gate traversal experiments further validate FoDeGlider's ability to actively reconfigure its morphology during locomotion, enabling enhanced navigation in confined underwater environments.
	\end{abstract}
	
	
	\section{Introduction}
	\IEEEPARstart{M}{iniature} hybrid underwater gliders combine the long endurance of buoyancy-driven vehicles with the maneuverability provided by propeller-assisted propulsion. This unique capability has attracted sustained interest in ocean inspection, environmental monitoring, and long-duration underwater missions \cite{wang2024motion,yang2022new}. However, despite recent advances, miniature hybrid underwater gliders still suffer limited maneuverability, particularly in confined environments that require enhanced traversability.
	To address these limitations, various designs have been investigated. Alternative propulsion and control mechanisms, including
	fish-like propulsors \cite{zhong2023design,zhang2013miniature} and controllable lifting surfaces \cite{han2024design}, were introduced to enhance control authority. Reconfigurable body structures, such as variable-area tails \cite{wang2024design}, movable bows \cite{wang2024motion}, foldable propulsion mechanisms \cite{chen2016design}, and volume-changing hulls \cite{ranganathan2018design}, were explored to improve maneuverability and operational versatility. 
    
    Among these approaches, geometry morphing is particularly attractive as it directly reconfigures the vehicle's hydrodynamic layout and thus alters the generated hydrodynamic forces and moments. In particular, wing morphing has received increasing attention because the wing is the primary lifting surface governing lift, drag, and pitch moment generation \cite{han2024design,wu2025design,wu2023multi}. By modifying wing geometry, morphing mechanisms reshape the lift--drag characteristics and broaden the achievable gliding performance envelope \cite{lentink2007swifts}. Existing studies have explored several wing-morphing strategies, including sweep-angle variation \cite{li2018variable,wang2022glide,zhang2023controllable}, angle-of-attack adjustment~ \cite{arima2009modelling}, and symmetric sweep morphing~\cite{rockenbauer2021dipper}.
    Despite these advances, existing wing-morphing designs typically support only a single dominant reconfiguration mode or a limited geometric range. As a result, they cannot simultaneously provide the large-range wing folding required for enhanced traversability and the independent wing deflection needed for agile maneuvering. 
	
	From a modeling perspective, most existing approaches adopt a single-rigid-body formulation in which hydrodynamic coefficients are parameterized as functions of sweep angle and wingspan under limited morphing ranges \cite{wang2017design,han2024design}. While effective for moderate geometric variation, such formulations fail to decouple the distinct hydrodynamic characteristics of fuselage and wings when large-range morphing severely alters the vehicle's effective frontal area, span, and center of pressure. 
    Furthermore, while component-based build-up methods aggregate loads from individual geometric elements \cite{fossen2011handbook,bhat2021real,bulka2018autonomous,wang2023dynamics}, they have not been extended to platforms where large-range wing motion introduces pronounced multibody effects---including configuration-dependent mass distribution, inertia coupling, and wrench transport. Therefore, a unified framework that integrates configuration-dependent multibody dynamics with component-level hydrodynamic modeling remains an open problem.

	To address the aforementioned challenges, this paper presents FoDeGlider, a  miniature hybrid underwater glider featuring two independently actuated foldable and deflectable wings that enable large-range shape reconfiguration while maintaining a compact fuselage architecture. To accurately capture the configuration-dependent dynamics induced by wing morphing, a multibody dynamics model is developed using spatial-operator formulations and composite rigid body algorithm (CRBA)-based projection. 
	To the best of the authors' knowledge, this work represents the first effort toward the development and multibody dynamics modeling of a miniature underwater glider with independently actuated large-range foldable and deflectable wings. The contributions of this paper are threefold.
	\begin{enumerate}
		\item A custom-designed miniature hybrid underwater glider, FoDeGlider, is developed with two independently actuated morphing wings. Each wing integrates a large-range folding mechanism ($0^\circ$--$90^\circ$ sweep) with an independent deflection mechanism ($0^\circ$--$90^\circ$). This integrated morphing architecture simultaneously improves geometric adaptability and maneuverability without increasing the vehicle footprint.
		\item A configuration-dependent multibody dynamics model is established for the morphing-wing underwater glider. Unlike conventional single-rigid-body glider models, the model treats wing configuration as a structural variable in the governing dynamics. It parameterizes the composite inertia, velocity transport, wrench transport, restoring wrench, and component hydrodynamics by the wing configuration through a CRBA-based projection, and assembles them into a unified Fossen-form equation.
		\item Extensive experiments are conducted under multiple sweep and deflection configurations to build an open-source dataset for parameter identification and model evaluation\footnote{All datasets, hardware designs, and modeling/system-identification codes are openly available at \url{https://github.com/zhu-yj/FoDeGlider}.}. A sequential parameter-identification framework is formulated for the fuselage and wing hydrodynamic models. The experiments validate the identified configuration-dependent model across multiple wing configurations.
	\end{enumerate}

	\section{System Design and Prototype Development}
	\label{sec:system_overview}
\begin{figure}[!t]
		\centering
		\includegraphics[width=\columnwidth]{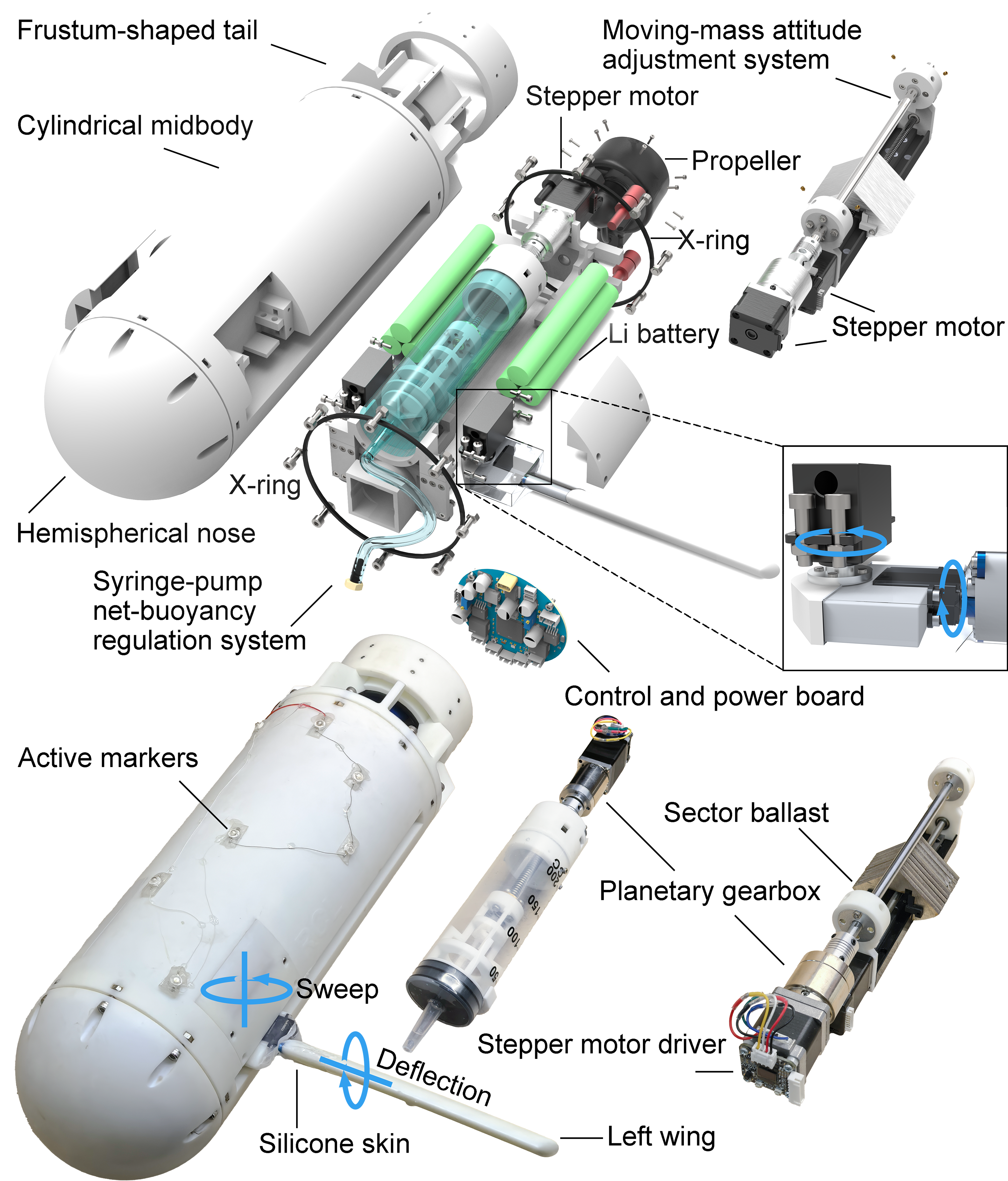}
		\caption{Design overview of FoDeGlider, a custom-designed miniature hybrid underwater glider featuring independently actuated large-range foldable and deflectable wings.}
		\label{fig:design}
	\end{figure}

	FoDeGlider, as illustrated in Fig.~\ref{fig:design}, represents a custom-designed prototype with independently actuated foldable and deflectable wings. The glider comprises a watertight fuselage, two morphing wing modules, a syringe-pump net-buoyancy regulation system, a moving-mass attitude adjustment system, an aft propeller, and an integrated control and power board.
	The prototype has a 513~mm overall length and a 6.71~kg dry mass, close to its 6.75~kg displaced-volume buoyancy.

	The additively manufactured fuselage consists of a hemispherical nose, a cylindrical midbody, and a frustum-shaped tail, axially sealed by nitrile rubber quad-rings. 
	Fuselage-embedded wing slots house two independently actuated, nearly neutrally buoyant hollow wings (119~g each). Each wing module features a root-mounted 30~kgf$\cdot$cm waterproof servo driving sweep over $[0^\circ,90^\circ]$, and a bracket-coupled 5~kgf$\cdot$cm micro servo actuating deflection over the same angular range.	
	A 0.2-mm-thick silicone skin encapsulates the deflection joint to preserve waterproofing.

\begin{figure}[!htbp]
		\centering
		\includegraphics[width=\columnwidth]{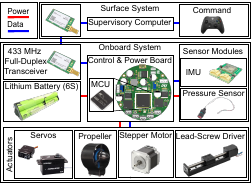}
		\caption{Embedded system architecture of FoDeGlider, including onboard electronics, communication, and power distribution.}
		\label{fig:elec}
	\end{figure}

	Net buoyancy is regulated by a syringe-pump system (100~mL maximum displacement) driven by a 28~mm stepper motor, changing the net buoyancy from $+0.04$~kg to $-0.06$~kg. Attitude adjustment is managed by a translating mass (100~mm range) for pitch control, and a rotating mass ($\pm 26^\circ$ range) for roll moments. The onboard STM32F407 microcontroller (MCU) processes depth and IMU sensor measurements and communicates with a surface computer through a 433~MHz full-duplex transceiver, while custom DC--DC converters manage power distribution. Fig.~\ref{fig:elec} summarizes the embedded electronics, communication links, and power-distribution architecture of the prototype.

	\section{Configuration-Dependent Kinematics and Dynamics Modeling}
	\label{sec:modeling}
	
	FoDeGlider is a kinematic tree of rigid bodies connected by revolute and prismatic joints, comprising the fuselage, two wing links, and three internal moving masses, namely a rotating ballast mass, a translating ballast mass, and a variable ballast mass. Its dynamics couple the motion of these bodies with the fluid loads acting on each body. To express these couplings in a coordinate-consistent form, we use spatial operators on the special Euclidean group $\SE$ \cite{marsden1999introduction,featherstone2008rigid}. This section first recalls the spatial-operator and composite-inertia preliminaries, then introduces the coordinate frames and configuration variables, and finally develops the configuration-dependent kinematic and dynamic models.
	
	\subsection{Preliminaries}
	\label{sec:preliminaries}
	In this work, a twist and its dual wrench are defined as $\bm{\nu}\triangleq[\bm{v}^T,\bm{\omega}^T]^T$ and $\bm{\tau}\triangleq[\bm{f}^T,\bm{m}^T]^T$, respectively. The relative pose between two coordinate frames $i$ and $j$ is an element $\bm{g}_{ij}\in\SE$ that combines the rotation $\prescript{i}{j}{\bm{R}}$ and the translation $\prescript{i}{}{\bm{r}}_{ij}$. The group adjoint operator $\Ad_{\bm{g}_{ij}}$ transforms twists between the frames, and the inverse-transpose adjoint transforms the dual wrenches. The wrench transformation is denoted by
	\begin{align}
		\prescript{i}{j}{\bm{X}}
		\triangleq
		\Ad_{\bm{g}_{ij}}^{-T}
		=
		\begin{bmatrix}
			\prescript{i}{j}{\bm{R}} & \zero_{3\times 3}\\
			[\prescript{i}{}{\bm{r}}_{ij}]^\times \prescript{i}{j}{\bm{R}} & \prescript{i}{j}{\bm{R}}
		\end{bmatrix},
		\label{eq:X}
	\end{align}
	where $[\cdot]^\times$ denotes the skew-symmetric matrix associated with the vector cross product. Thus, a wrench maps as $\bm{\tau}_i=\prescript{i}{j}{\bm{X}}\bm{\tau}_j$ and a twist maps as $\bm{\nu}_j=\prescript{i}{j}{\bm{X}}^T\bm{\nu}_i$. The Lie algebra adjoint operator $\adop$ and its dual $\adop^*$ are written consistently with these definitions.

	With these operators, the momentum balance of a rigid body is written in the forced Euler--Poincar\'{e} form \cite{marsden1999introduction,moghaddam2023singularity}, i.e.,
	\begin{align} 
		\bm{M}\dot{\bm{\nu}} - \adop_{\bm{\nu}}^*(\bm{M}\bm{\nu}) = \bm{\tau},
		\label{eq:ep_local}
	\end{align}
	where $\bm{M}$ denotes the spatial inertia matrix. $\adop_{\bm{\nu}}^*(\bm{M}\bm{\nu})$ carries the Coriolis and centripetal terms and, through the added inertia, the Munk-type contribution. Equation~\eqref{eq:ep_local} is applied to each body and projected to frame $i$ to assemble the multibody model. 
	
	The spatial inertia in \eqref{eq:ep_local} maps a twist to spatial momentum, given by
	\begin{align}
		\bm{M}
		=
		\begin{bmatrix}
			m\I_3 & -m[\bm{r}_{\mathrm{CM}}]^\times\\
			m[\bm{r}_{\mathrm{CM}}]^\times & \bm{I}_{\mathrm{CM}} - m[\bm{r}_{\mathrm{CM}}]^\times[\bm{r}_{\mathrm{CM}}]^\times
		\end{bmatrix},
		\label{eq:spatial_inertia}
	\end{align}
	where $m$ is the mass, $\bm{r}_{\mathrm{CM}}$ is the center-of-mass offset from the frame origin, and $\bm{I}_{\mathrm{CM}}$ is the rotational inertia about the center of mass. A point mass is the degenerate case $\bm{I}_{\mathrm{CM}}=\zero$, and for a submerged body $\bm{M}$ additionally absorbs the hydrodynamic added inertia, so the same algebra applies to the combined inertia. When the joints are locked at the current condition, the twist of every body is a linear function of the twist $\bm{\nu}_{i}$, and momentum projects to frame $i$ in the same way as a wrench. Summing the projected body inertias yields the configuration-dependent composite inertia computed by CRBA \cite{featherstone2008rigid}, which is expressed as $\sum_j \prescript{i}{j}{\bm{X}}\bm{M}_j\prescript{i}{j}{\bm{X}}^T$.

	\subsection{Coordinate Frames and Configuration Variables}
	\label{sec:frames}
\begin{figure}[!t]
		\centering
		\includegraphics[width=\columnwidth]{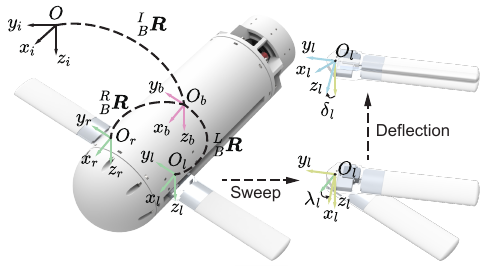}
		\caption{Coordinate frames defined for the multibody model of FoDeGlider. The inertial frame $I$, base frame $B$ attached to the fuselage, and wing frames $L$ and $R$ are introduced, with the left-wing frame transformation governed by the sweep and deflection configurations.}
		\label{fig:frame}
	\end{figure}

	Let $B$ denote the base frame attached to the fuselage and $I$ the inertial frame. The vehicle comprises two morphing wing links forming the wing set $\mathcal{W}=\{l,r\}$ and three internal actuation bodies $\{p_1,p_2,p_3\}$, namely the rotating ballast mass, the translating ballast mass, and the variable ballast mass. Together they form the movable-body set $\mathcal{B}=\{l,r,p_1,p_2,p_3\}$. The fuselage is treated separately from $\mathcal{B}$ and is assigned the subscript $b$, with $B$ denoting the base frame attached to it. Frames are denoted by uppercase letters and bodies by lowercase subscripts. A leading superscript specifies the frame in which a quantity is expressed, as in $\prescript{B}{}{\bm{r}}$. A double decoration specifies a transformation, as in $\prescript{B}{L}{\bm{R}}$, which maps vectors from frame $L$ to frame $B$. A right subscript represents the body associated with a quantity, and quantities of the fuselage use the subscript $b$. The coordinate frames are illustrated in Fig.~\ref{fig:frame}.

	The unified joint-coordinate vector collects the morphing-wing and internal-actuation degrees of freedom as
	\begin{align}
		\bm{q}_j =
		[\lambda_l,\delta_l,\lambda_r,\delta_r,\theta_m,d_m,d_p]^T
		\in \mathbb{R}^7 
		\label{eq:qj}
	\end{align}
	where $[\lambda_l,\delta_l,\lambda_r,\delta_r]$ are defined as configuration variables, denoting the sweep and deflection angles of the left wing and right wing, respectively. The internal joint-coordinate vector is defined as $\bm{q}_{j,\mathrm{int}}=[\theta_m,d_m,d_p]^T$, where $\theta_m$, $d_m$, and $d_p$ denote the rotating-ballast angle, the translating-ballast displacement along the longitudinal body axis, and the piston displacement, respectively. 

	\subsection{Configuration-based Kinematics}
	\label{sec:base_kin}

	The pose of the fuselage relative to the inertial frame is $\bm{\eta}=[\bm{p}^T\ \bm{e}^T]^T=[x,y,z,\phi,\theta,\psi]^T$, where $\bm{p}\in\mathbb{R}^3$ is the position of the fuselage expressed in $I$, and $\bm{e}=[\phi,\theta,\psi]^T$ is the ZYX Euler-angle vector. The twist of the fuselage is defined as $\bm{\nu}_b=[\bm{v}_b^T\ \bm{\omega}_b^T]^T\in\R^6$, where $\bm{v}_b$ and $\bm{\omega}_b$ are the linear and angular velocities of the fuselage expressed in $B$. For notational compactness, the leading superscript $B$ which represents the base frame is omitted here.

    The base twist maps to the inertial pose rate through
	\begin{align}
		\dot{\bm{\eta}} = \bm{J}_{\eta}(\bm{e})\,\bm{\nu}_b,
		\qquad
		\bm{J}_{\eta}(\bm{e}) =
		\begin{bmatrix}
			\prescript{I}{B}{\bm{R}}(\bm{e}) & \zero_{3\times3}\\
			\zero_{3\times3} & \bm{T}_{\eta}(\bm{e})
		\end{bmatrix},
		\label{eq:kinematics}
	\end{align}
	where $\bm{T}_{\eta}(\bm{e})$ is the ZYX Euler-angle rate transformation matrix, which maps the body angular velocity to the Euler-angle rates. Equation~\eqref{eq:kinematics} provides the mapping used in the experimental analysis to obtain the linear and angular velocities expressed in the base frame, together with their accelerations, from the measured inertial pose trajectory.

	For wing body $k\in\{l,r\}$ with wing-fixed frame $K\in\{L,R\}$, the origin of $K$ is placed at the wing hinge $D_k$, so the relative wing motion has no translational part in the joint-motion subspace. The orientation of $K$ relative to the base frame $B$ is given by
	\begin{align}
		\prescript{B}{K}{\bm{R}}
		=
		\prescript{B}{K_1}{\bm{R}}(\lambda_k)\,
		\prescript{K_1}{K}{\bm{R}}(\delta_k),
		\label{eq:Rk}
	\end{align}
	where $K_1$ is the intermediate frame after the sweep rotation. This rotation and the hinge position define the wing pose $\bm{g}_{BK}\in\SE$ relative to the base frame. The corresponding wrench transformation is $\prescript{B}{K}{\bm{X}}=\Ad_{\bm{g}_{BK}}^{-T}$, following the corresponding form in~\eqref{eq:X}.
	
\begin{figure}[!t]
		\centering
		\includegraphics[width=\columnwidth]{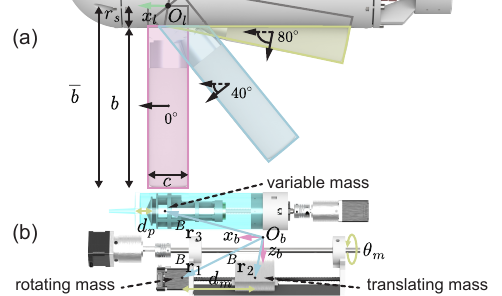}
		\caption{Morphing-wing and internal moving-mass geometry of FoDeGlider. (a)~Wing-fixed frame and effective wing area under representative sweep angles $0^\circ$, $40^\circ$, and $80^\circ$. (b)~Internal moving-mass geometry, illustrating the rotating mass, translating mass, and variable mass, together with their center-of-mass locations.}
		\label{fig:vector}
	\end{figure}

	The joint-motion subspace matrix that maps wing joint rates to the corresponding twist is
	\begin{align}
		\bm{J}_k =
		\begin{bmatrix}
			\zero_{3\times 2}^T &
			\bm{J}_{\omega k}^T
		\end{bmatrix}^T,
		\quad
		\bm{J}_{\omega k} =
		\begin{bmatrix}
			\prescript{K}{B}{\bm{R}}\prescript{B}{}{\bm{e}}_z &
			\prescript{K}{}{\bm{e}}_y
		\end{bmatrix},
		\label{eq:Jk}
	\end{align}
	where $\prescript{B}{}{\bm{e}}_z$ is the unit vector along the base-frame $z$ axis, which defines the sweep axis. The vector $\prescript{K}{}{\bm{e}}_y$ is the unit vector along the wing-frame $y$ axis, which defines the deflection axis.
	Hence the twist of wing body $k$, expressed in frame $K$, is given by
	\begin{align}
		\prescript{K}{}{\bm{\nu}}_k
		=
		\prescript{B}{K}{\bm{X}}^T\bm{\nu}_b
		+ \bm{J}_k\dot{\bm{q}}_{j,w,k},
		\qquad
		\bm{q}_{j,w,k} = [\lambda_k,\delta_k]^T,
		\label{eq:nuk}
	\end{align}
	where $\prescript{B}{K}{\bm{X}}^T\bm{\nu}_b$ transports the fuselage velocity expressed in the base frame to the wing frame, and $\bm{J}_k\dot{\bm{q}}_{j,w,k}$ represents the relative motion induced by the wing joints. The wing-fixed frame and the effective area at representative sweep angles are shown in Fig.~\ref{fig:vector}(a).

    As illustrated in Fig.~\ref{fig:vector}(b), the stepper motor and slide assembly constitute the rotating mass $p_1$, whose motion is parameterized by $\theta_m$. The mass block on the slide defines the translating mass $p_2$, which undergoes coupled rotational motion with $p_1$ and translational motion along the slide described by $d_m$. The water and piston in the syringe-pump system form the variable mass $p_3$, whose position varies with the piston displacement $d_p$. Accordingly, the center-of-mass positions of these three components expressed in the base frame are formulated as $\prescript{B}{}{\bm{r}}_1(\theta_m)$, $\prescript{B}{}{\bm{r}}_2(\theta_m,d_m)$, and $\prescript{B}{}{\bm{r}}_3(d_p)$. 
	Their point-mass spatial inertias follow from \eqref{eq:spatial_inertia} with $\bm{I}_{\mathrm{CM}}=\zero$ evaluated at $\prescript{B}{}{\bm{r}}_i$, and their twists are
	\begin{align}
		\bm{\nu}_{p,i}
		=
		\bm{\nu}_b
		+
		\bm{J}_{p,i}\dot{\bm{q}}_{j,\mathrm{int}},
		\qquad
		\bm{J}_{p,i}
		=
		\begin{bmatrix}
			\bm{J}_{m,i}^T & \zero_{3\times 3}^T
		\end{bmatrix}^T,
		\label{eq:nupi}
	\end{align}
	where $\bm{J}_{m,i}\in\R^{3\times3}$ is the translational Jacobian of point mass $p_i$ with respect to $\bm{q}_{j,\mathrm{int}}$. The internal-mass layout is illustrated in Fig.~\ref{fig:vector}(b).		
	
	\subsection{Configuration-based Dynamics}
	\label{sec:eom}
	Applying the Euler--Poincar\'{e} primitive~\eqref{eq:ep_local} to each body and projecting to the base frame through the composite-inertia construction in Section~\ref{sec:preliminaries} yields the configuration-dependent governing equation in the Fossen form
	\begin{align}
		& \bm{M}(\bm{q}_j)\dot{\bm{\nu}}_b
		+ \bm{C}(\bm{q}_j,\bm{\nu}_b)\bm{\nu}_b
		+ \bm{M}_{bj}(\bm{q}_j)\ddot{\bm{q}}_j \notag \\
		& \quad
		+ \bm{b}_j(\bm{q}_j,\dot{\bm{q}}_j,\bm{\nu}_b)
		+ \bm{g}(\bm{q}_j,\bm{\eta})
		=
		\bm{\tau}_{\mathrm{ext}}.
		\label{eq:eom}
	\end{align}
	The first two terms represent the composite inertia and the locked-joint velocity-product effects, $\bm{M}_{bj}(\bm{q}_j)\ddot{\bm{q}}_j$ is the inertia coupling caused by prescribed joint accelerations, $\bm{b}_j$ collects the remaining velocity coupling induced by nonzero joint rates, $\bm{g}(\bm{q}_j,\bm{\eta})$ is the gravity--buoyancy restoring wrench, and $\bm{\tau}_{\mathrm{ext}}$ is the external wrench developed in Section~\ref{sec:hydro}. Each movable body is projected to the base frame through the wrench transformation
	\begin{align}
		\prescript{B}{K}{\bm{X}} =
		\begin{cases}
			\Ad_{\bm{g}_{BK}}^{-T}, & k\in\mathcal{W},\\
			\I_6, & k\in\mathcal{B}\setminus\mathcal{W},
		\end{cases}
		\label{eq:Pkappa}
	\end{align}
	where the identity applies to the internal masses, whose twists are already expressed in the base frame.
	
	Specializing the composite-inertia projection to the prototype gives
	\begin{align}
		\bm{M}(\bm{q}_j)
		=
		\bm{M}_b
		+
		\sum_{k\in\mathcal{B}}
		\prescript{B}{K}{\bm{X}}
		\bm{M}_k \prescript{B}{K}{\bm{X}}^T,
		\label{eq:M}
	\end{align}
	where the fuselage spatial inertia $\bm{M}_b$ is expressed at the base-frame origin and therefore enters \eqref{eq:M} without transformation. $\bm{M}_k$ is the spatial inertia of movable body $k$ in its own frame $K$. This construction preserves the symmetry and positive definiteness of $\bm{M}(\bm{q}_j)$. For each wing body $k\in\mathcal{W}$, $\bm{M}_k$ is referred to the hinge $D_k$ and combines a rigid-body and a configuration-dependent added-mass part,
	\begin{align}
		\bm{M}_k^{D}
		=
		\bm{M}_{RB,k}^{D}
		+
		\bm{M}_{A,k}^{D}.
		\label{eq:MkD}
	\end{align}
	The rigid-body part $\bm{M}_{RB,k}^{D}$ is the spatial inertia in~\eqref{eq:spatial_inertia} evaluated at the hinge $D_k$ by the parallel-axis theorem. The added-mass part is first expressed at the geometric center of the effective wing area, i.e.,
	\begin{align}
		\bm{M}_{A,k}^{\mathrm{gc}}
		=
		\begin{bmatrix}
			\bm{M}_{ak}^{\mathrm{eff}}(\lambda_k) & \zero_{3\times3}\\
			\zero_{3\times3} & \bm{I}_{ak}^{\mathrm{eff}}(\lambda_k)
		\end{bmatrix},
		\label{eq:MAH}
	\end{align}
	where the blocks $\bm{M}_{ak}^{\mathrm{eff}}$ and $\bm{I}_{ak}^{\mathrm{eff}}$ are defined in Section~\ref{sec:hydro}. This block is then transported from the geometric center to the hinge $D_k$ by the parallel-axis theorem to give $\bm{M}_{A,k}^{D}$. Because the geometric center $\prescript{K}{}{\bm{r}}_{\mathrm{gc}}(\lambda_k)$ migrates with the sweep angle, this keeps the added mass referred to a consistent hinge point as the wing folds.

	The Coriolis and centripetal matrix $\bm{C}$ is built through the same base-frame projection used for the inertia in~\eqref{eq:M}. For body $k$ at its locked-joint velocity $\bm{\nu}_{k,0}=\prescript{B}{K}{\bm{X}}^T\bm{\nu}_b$, the Coriolis matrix takes the Fossen form~\cite{fossen2011handbook}, and projecting it to the base frame gives
	\begin{align}
		\bm{C}(\bm{q}_j,\bm{\nu}_b)
		=
		\bm{C}_b(\bm{\nu}_b)
		+
		\sum_{k\in\mathcal{B}}
		\prescript{B}{K}{\bm{X}}\,
		\bm{C}_k(\bm{\nu}_{k,0})\,
		\prescript{B}{K}{\bm{X}}^T,
		\label{eq:Cnu}
	\end{align}
	where the matrix $\bm{C}_k(\bm{\nu}_{k,0})$ is derived from the spatial-inertia blocks as~\cite{fossen2011handbook}

	\begingroup
	\small
	\setlength{\arraycolsep}{2pt}
	\renewcommand{\arraystretch}{0.88}
	\begin{equation}
	\begin{bmatrix}
	\zero_{3\times3} &
	-[\bm{M}_{k,vv}\bm{v}_{k,0}
	+\bm{M}_{k,v\omega}\bm{\omega}_{k,0}]^\times \\
	-[\bm{M}_{k,vv}\bm{v}_{k,0}
	+\bm{M}_{k,v\omega}\bm{\omega}_{k,0}]^\times &
	-[\bm{M}_{k,\omega v}\bm{v}_{k,0}
	+\bm{M}_{k,\omega\omega}\bm{\omega}_{k,0}]^\times
	\end{bmatrix}
	\notag
	\end{equation}
	\endgroup

	The joint-acceleration coupling and the bias force are
	\begin{align}
		& \bm{M}_{bj}\ddot{\bm{q}}_j
		=
		\sum_{k\in\mathcal{B}}
		\prescript{B}{K}{\bm{X}}\,\bm{M}_k\bm{J}_k\ddot{\bm{q}}_{j,k}, \text{ and}
		\notag \\
		& \bm{b}_j
		=
		\sum_{k\in\mathcal{B}}
		\prescript{B}{K}{\bm{X}}
		\big[
		\bm{M}_k\bm{\sigma}_k
		-\adop_{\bm{\nu}_k}^{*}(\bm{M}_k\bm{\nu}_k)
		+\adop_{\bm{\nu}_{k,0}}^{*}(\bm{M}_k\bm{\nu}_{k,0})
		\big] \notag
	\end{align}
	where $\bm{q}_{j,k}$ is the joint subset of body $k$ and $\bm{\sigma}_k=\dot{\bm{J}}_k\dot{\bm{q}}_{j,k}+\adop_{\bm{\nu}_k}(\bm{J}_k\dot{\bm{q}}_{j,k})$ is the joint-induced velocity-product bias. The term $\bm{C}\bm{\nu}_b$ thus carries the locked-joint velocity-product effects, while $\bm{b}_j$ accounts for the additional bias generated by prescribed joint motion.

	The restoring wrench is assembled by the corresponding wrench projection, i.e.,
	\begin{align}
		\bm{g}(\bm{q}_j,\bm{\eta})
		=
		\bm{g}_b
		+
		\sum_{k\in\mathcal{B}}
		\prescript{B}{K}{\bm{X}}\,\bm{g}_k,
		\label{eq:g}
	\end{align}
	where $\prescript{B}{K}{\bm{X}}=\I_6$ for the internal point masses. The restoring wrench of the fuselage is
	\begin{align}
		\bm{g}_b =
		\begin{bmatrix}
			-(m_b g-B_b)\,\prescript{B}{}{\bm{e}}_g\\
			-(m_b g\,\prescript{B}{}{\bm{r}}_{b,\mathrm{CM}}
			-B_b\,\prescript{B}{}{\bm{r}}_{b,\mathrm{CB}})
			\times\prescript{B}{}{\bm{e}}_g
		\end{bmatrix},
		\label{eq:gb}
	\end{align}
	where $\prescript{B}{}{\bm{e}}_g=\prescript{B}{I}{\bm{R}}[0,0,1]^T$ is the vertical unit vector in the base frame, $m_b$ is the fuselage mass, and $B_b$ its buoyancy. The wing restoring wrench $\bm{g}_k$ has the identical structure, referred to the hinge $D_k$ and expressed in frame $K$ with $\prescript{K}{}{\bm{e}}_g=\prescript{K}{B}{\bm{R}}\,\prescript{B}{}{\bm{e}}_g$, retaining both the gravity and the buoyancy terms. The internal masses are non-buoyant, so their restoring wrench keeps only the gravitational force $-m_i g\,\prescript{B}{}{\bm{e}}_g$ and its moment $-m_i g\,[\prescript{B}{}{\bm{r}}_i]^\times\prescript{B}{}{\bm{e}}_g$.

	\section{Configuration-Dependent Hydrodynamic Modeling}
	\label{sec:hydro}
	
	Large-range wing morphing introduces strong configuration-dependent hydrodynamic variations. To capture this effect, a component-wise hybrid model is adopted. The fuselage is represented using low-order base-frame damping and treated independently from morphing-induced wing effects. Each wing is modeled using trigonometric functions of local angle of attack and sideslip angle. The formulation accounts for variations in effective area, span, chord, center of geometry, and added-mass through parameterized force and moment coefficients. The total hydrodynamic wrench is obtained by decoupling fuselage viscous dynamics and wing morphing effects prior to assembly in the base-frame dynamics.

	\subsection{Sweeping Wing Geometry and Scaling Function Design}
	Let $\lambda \triangleq |\lambda_k|$ denote the wing sweep magnitude, which determines the effective wing geometry. Let $b$ and $c$ be the nominal effective span and chord at $\lambda=0$, and let $r_s$ denote the wing-root offset, i.e., the spanwise distance from the sweep axis to the fuselage surface. Defining $\bar b=b+r_s$ as the distance from the sweep axis to the wingtip, the effective wing area outside the fuselage boundary is modeled as
	\begin{align}
		S^{\mathrm{eff}}(\lambda) =
		\left\{
        \begin{array}{@{}l@{\,}l@{}}
			c\left(\bar b-\dfrac{r_s}{\cos\lambda}\right), & 0\leq \lambda \leq \lambda_c\\[8pt]
			\dfrac{\tan\lambda}{2}\left(\dfrac{\bar b\cos\lambda-r_s}{\sin\lambda}+\dfrac{c}{2}\right)^2,
			& \lambda_c < \lambda < 90^\circ
		\end{array}
        \right.
		\label{eq:Sexp}
	\end{align}
	where the angle $\lambda_c$ is calculated as
	\begin{align}
		\lambda_c = \arcsin\left(\frac{(\bar b/r_s)^2-1}{(\bar b/r_s)^2+1}\right).
		\label{eq:thetac}
	\end{align}
	Equation~\eqref{eq:Sexp} explicitly tracks the effective wing area, which affects both hydrodynamics and added mass. Computational studies on swift-inspired swept wings have similarly shown that
	sweep variation changes the effective wing geometry, especially the wing area and aspect ratio, and thereby affects the drag, lift, and hydrodynamic moment coefficients of underwater gliders \cite{wang2022glide}.
	
	The geometric center of the effective area is used as the reference point for wing loads with its wing-frame coordinates
		\begin{align}
		x_{\mathrm{gc}}(\lambda) &=
		\left\{
        \begin{array}{@{}l@{\,}l@{}}
		-\dfrac{c^2\tan\lambda}{12(\bar b-r_s/\cos\lambda)}, 
		& 0\leq \lambda \leq \lambda_c\\
		\dfrac{1}{3}\left(\dfrac{\bar b\cos\lambda-r_s}{\sin\lambda}-c\right), 
		& \lambda_c < \lambda \leq 90^\circ
		\end{array}
        \right.
        \notag
		\\
		y_{\mathrm{gc}}(\lambda) &=
		\left\{
        \begin{array}{@{}l@{\,}l@{}}
		\dfrac{1}{2}\left(\bar b+\dfrac{r_s}{\cos\lambda}\right)
		-\dfrac{c^2\tan^2\lambda}{24(\bar b-r_s/\cos\lambda)}, 
		& \; 0\leq \lambda \leq \lambda_c\\
		\dfrac{1}{3}\left(\dfrac{r_s-(c/2)\sin\lambda}{\cos\lambda}+2\bar b\right), 
		& \lambda_c < \lambda \leq 90^\circ
		\end{array}
        \right.
        \notag
		\end{align}

	The corresponding geometric center vectors for the left and right wings are
	\begin{align}
		\prescript{L}{}{\bm{r}}_{\mathrm{gc}}(\lambda_l)
		=
		\begin{bmatrix}
			-x_{\mathrm{gc}},
			-y_{\mathrm{gc}},
			0
		\end{bmatrix}^T,
		\;
		\prescript{R}{}{\bm{r}}_{\mathrm{gc}}(\lambda_r)
		=
		\begin{bmatrix}
			-x_{\mathrm{gc}},
			y_{\mathrm{gc}},
			0
		\end{bmatrix}^T
		\label{eq:rcp}
	\end{align}
	
	The characteristic lengths used in the moment terms are the effective span projection and the area-equivalent chord, i.e.,
	\begin{align}
		b_k^{\mathrm{eff}}(\lambda)
		&=
		\bar b\cos\lambda+\frac{c}{2}\sin\lambda-\frac{c}{2},
		&
		c_k^{\mathrm{eff}}(\lambda)
		&=
		\frac{S^{\mathrm{eff}}(\lambda)}{b_k^{\mathrm{eff}}(\lambda)}.
        \notag
	\end{align}
	Here, the quantity $b_k^{\mathrm{eff}}$ provides the effective length scale for roll and yaw moments, whereas $c_k^{\mathrm{eff}}$ provides the effective length scale for pitch moments. Near the fully folded configuration, both $S^{\mathrm{eff}}$ and $b_k^{\mathrm{eff}}$ approach zero, whereas their ratio remains finite. The corresponding hydrodynamic force and moment terms also tend to zero because they are scaled by the dynamic pressure-area term defined below.

	The normalized geometry ratios used for coefficient and added-mass scaling are
	\begin{align}
		\eta_S=\frac{S^{\mathrm{eff}}}{S_0},\quad
		\eta_b=\frac{b_k^{\mathrm{eff}}}{b_0},\quad
		\eta_I=\eta_S\eta_b^2,
		\qquad S_0=bc .
		\label{eq:scale_inputs}
	\end{align}

	To model the remaining configuration dependence while enforcing the prescribed endpoint constraints, two positive quadratic Bernstein scaling functions are introduced
	\begin{align}
		K_{\mathrm{H}}(\eta;\zeta_{H,0},\zeta_{H,1})
		&=
		\zeta_{H,0}(1-\eta)^2
		+2\zeta_{H,1}\eta(1-\eta)
		+\eta^2,
        \notag
		\\
		K_{\mathrm{A}}(\eta;\zeta_{A})
		&=
		2\zeta_{A}\eta(1-\eta)+\eta^2.
		\label{eq:scale_added}
	\end{align}
	The function $K_{\mathrm{H}}$ scales the hydrodynamic coefficients and satisfies $K_{\mathrm{H}}(1;\zeta_{H,0},\zeta_{H,1})=1$. The function $K_{\mathrm{A}}$ scales the added-mass and added-inertia terms and satisfies $K_{\mathrm{A}}(0;\zeta_{A})=0$ and $K_{\mathrm{A}}(1;\zeta_{A})=1$. All scaling parameters are constrained to be strictly positive, ensuring positive hydrodynamic scaling and enforcing endpoint constraints.
    Constant-coefficient hydrodynamic models cannot capture the changes induced by wing morphing \cite{mclain1998development,wang2015averaging,han2024design}. To account for geometry-dependent hydrodynamic effects while maintaining model compactness, the normalized geometry ratios in \eqref{eq:scale_inputs} are adopted as scaling variables for both wing hydrodynamics and added-mass terms.

\begin{figure}[!htbp]		
		\centering
		\includegraphics[width=\columnwidth]{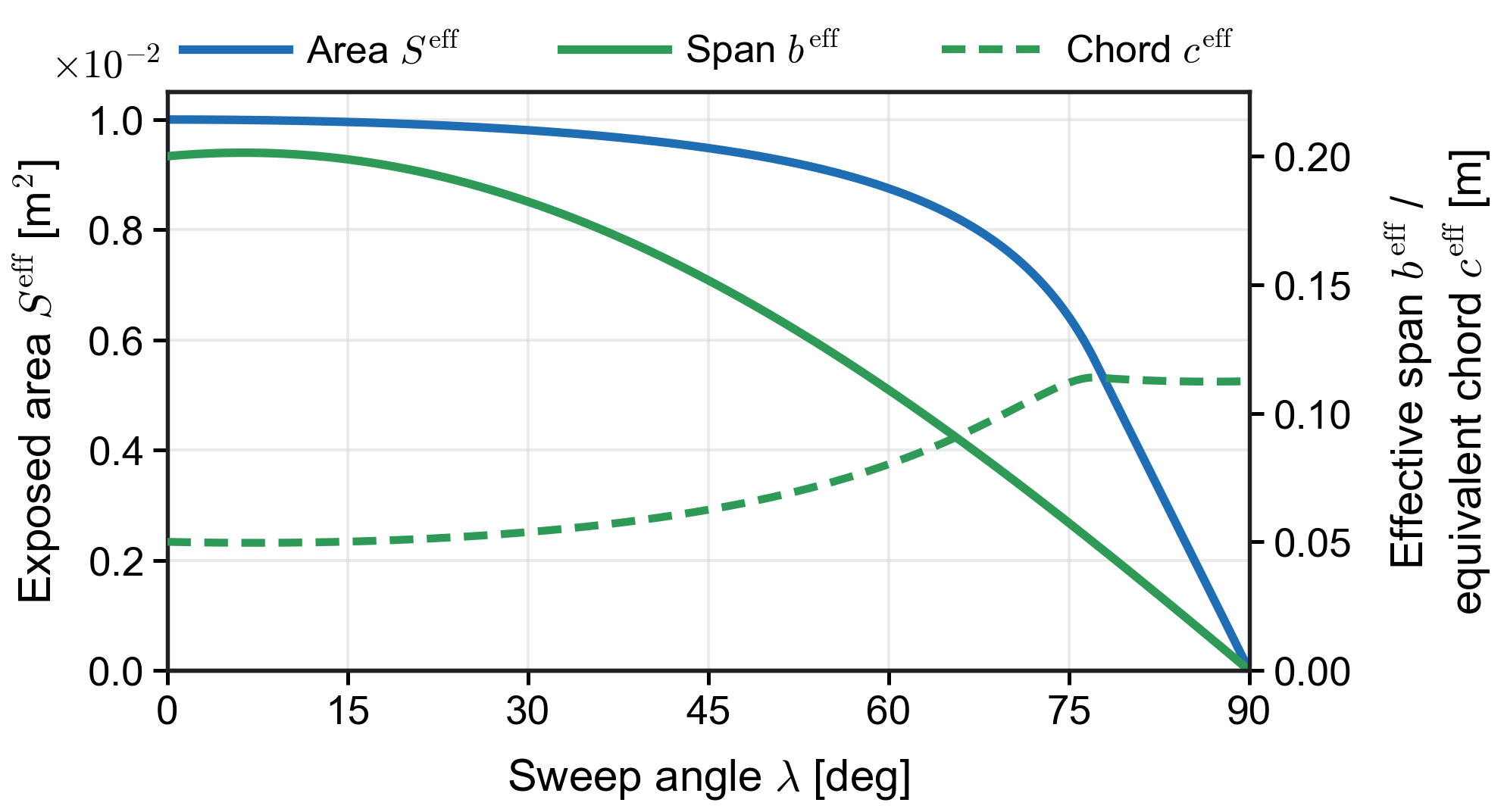}
		\caption{Representative wing morphing configurations and sweep-dependent geometric parameters for hydrodynamic modeling, including the effective area, effective span, and area-equivalent chord of FoDeGlider.}
		\label{fig:wingstate}
	\end{figure}

    Figure~\ref{fig:wingstate} illustrates the variation of effective area, effective span, and area-equivalent chord with sweep angle $\lambda$.
	
	\subsection{Wing Hydrodynamic Wrench Modeling}
    The effective added mass and added inertia of each wing are written as 
	\begin{align}
		\bm{M}_{ak}^{\mathrm{eff}}(\lambda_k)
		& =
		\diag\!\left(\bm{d}_{M,k}\right)\odot \bm{M}_{ak},\quad \\
		\bm{I}_{ak}^{\mathrm{eff}}(\lambda_k)
		& =
		\diag\!\left(\bm{d}_{I,k}\right)\odot \bm{I}_{ak}
		\label{eq:Iaeff}
	\end{align}
	where $\odot$ denotes the Hadamard product, and the scaling vectors are
	\begin{align}
		\bm{d}_{M,k}
		=&
		\Big[
		K_{\mathrm{A}}(\eta_S;\zeta_{A,X}),
		K_{\mathrm{A}}(\eta_S;\zeta_{A,Y}),
		K_{\mathrm{A}}(\eta_S;\zeta_{A,Z})
		\Big]^T,
		\notag
		\\
		\bm{d}_{I,k}
		=&
		\Big[
		K_{\mathrm{A}}(\eta_I;\zeta_{A,K}),
		K_{\mathrm{A}}(\eta_S;\zeta_{A,M}),
		K_{\mathrm{A}}(\eta_I;\zeta_{A,N})
		\Big]^T.
        \notag
	\end{align}
	Because $K_{\mathrm{A}}(0;\zeta)=0$, the added-mass and added-inertia contributions tend to zero as the effective wing area vanishes. The parameters $\zeta_{A,\cdot}$ control the nonlinear interpolation between this zero-area limit and the nominal value at $\eta=1$.

	The flow velocity at the geometric center of wing $k$ is
	\begin{align}
		\bm{v}_{\mathrm{f},k}
		=
		\bm{v}_k + \bm{\omega}_k\times\prescript{K}{}{\bm{r}}_{\mathrm{gc}}(\lambda_k)
		\label{eq:vflow}
	\end{align}
	The local angle of attack and sideslip angle are defined as
	\begin{align}
		\alpha_k = \arctan\!\left(\frac{v_{\mathrm{f},z}}{v_{\mathrm{f},x}}\right),
		\, \text{and }
		\beta_k = \arcsin\!\left(\frac{v_{\mathrm{f},y}}{V_{\mathrm{f},k}}\right),
		\label{eq:beta}
	\end{align}
	where $v_{\mathrm{f},x}$, $v_{\mathrm{f},y}$, and $v_{\mathrm{f},z}$ are the components of $\bm{v}_{\mathrm{f},k}$ in the wing frame, and $V_{\mathrm{f},k}=\|\bm{v}_{\mathrm{f},k}\|$ is its magnitude. 
    A trigonometric parameterization~\cite{bhat2021real,li2022centre} is adopted to model the configuration-dependent wing hydrodynamics over large sweep and deflection angles. This representation naturally captures angle-dependent variations while maintaining validity across a broad operating envelope.

	Defining the dynamic pressure-area factor as $q_{d,k}=\tfrac{1}{2}\rho V_{\mathrm{f},k}^2 S^{\mathrm{eff}}$, the wing force model is described as
	\begin{align}
		D_k
		&=
		q_{d,k}
		\left(
		C_{D0,k}^{\mathrm{eff}}
		+ C_{D\alpha,k}^{\mathrm{eff}}\sin^2\alpha_k
		\right),
		\label{eq:D}
		\\
		L_k
		&=
		q_{d,k}
		\left(
		C_{L\alpha,k}^{\mathrm{eff}}\sin\alpha_k\cos\alpha_k
		\right),
		\label{eq:L}
		\\
		S_k
		&=
		q_{d,k}
		\left(
		C_{S\beta,k}^{\mathrm{eff}}\sin\beta_k\cos\beta_k
		\right).
		\label{eq:S}
	\end{align}
	In \eqref{eq:D}--\eqref{eq:S}, $C_{D0,k}^{\mathrm{eff}}$ denotes the zero-incidence drag coefficient, $C_{D\alpha,k}^{\mathrm{eff}}$ denotes the angle-dependent drag coefficient, and $C_{L\alpha,k}^{\mathrm{eff}}$ and $C_{S\beta,k}^{\mathrm{eff}}$ denote the lift and side-force coefficients, respectively. These coefficients are defined with respect to the wing geometry induced by the sweep angle through the scaling functions introduced below.
	
	Using the wind-to-wing-axis transformation $\bm{R}_{wa}(\alpha_k,\beta_k)$, the hydrodynamic force on wing $k$ is
	\begin{align}
		\prescript{K}{}{\bm{f}}_{\mathrm{h,k}}
		=
		\bm{R}_{wa}(\alpha_k,\beta_k)
		\begin{bmatrix}
			-D_k &
			S_k &
			-L_k
		\end{bmatrix}^T
		\label{eq:Fhydro}
	\end{align}

	The nondimensional angular-rate terms are defined as
	\begin{align}
		p_k^\ast=\frac{p_k b_k^{\mathrm{eff}}}{2V_{\mathrm{f},k}},\quad
		q_k^\ast=\frac{q_k c_k^{\mathrm{eff}}}{2V_{\mathrm{f},k}},\, \text{and }
		r_k^\ast=\frac{r_k b_k^{\mathrm{eff}}}{2V_{\mathrm{f},k}},
		\label{eq:rate_norm}
	\end{align}
	where $\prescript{K}{}{\bm{\omega}}_k=[p_k,q_k,r_k]^T$ is the angular velocity of wing $k$ expressed in the wing frame. The quasi-steady moment terms are first computed in the local
	wind frame, i.e.,
	\begin{align}
		m_{x,k}
		&=
		q_{d,k} b^{\mathrm{eff}}_k
		C_{x\beta,k}^{\mathrm{eff}}\sin\beta_k\cos\beta_k,
		\label{eq:Mx_static}
		\\
		m_{y,k}
		&=
		q_{d,k} c^{\mathrm{eff}}_k
		\left(
		C_{m0,k}^{\mathrm{eff}}
		+ C_{m\alpha,k}^{\mathrm{eff}}\sin\alpha_k\cos\alpha_k
		\right), 
		\label{eq:My_static}
		\\
		m_{z,k}
		&=
		q_{d,k} b^{\mathrm{eff}}_k
		C_{z\beta,k}^{\mathrm{eff}}\sin\beta_k\cos\beta_k .
		\label{eq:Mz_static}
	\end{align}
	The damping moment is formulated directly in the wing frame
	\begin{align}
		\prescript{K}{}{\bm{m}}_{\mathrm{d},k}
		=
		-q_{d,k}\begin{bmatrix}
			 b_k^{\mathrm{eff}} C_{xp,k}^{\mathrm{eff}}p_k^\ast,
			c_k^{\mathrm{eff}} C_{mq,k}^{\mathrm{eff}}q_k^\ast,
			b_k^{\mathrm{eff}} C_{zr,k}^{\mathrm{eff}}r_k^\ast
		\end{bmatrix}^T
        \notag
	\end{align}
	Thus, the moment about the geometric center, expressed in the wing frame, is given by
	\begin{align}
		\prescript{	K}{}{\bm{m}}_{k}^{\mathrm{gc}}
		=
		\bm{R}_{wa}(\alpha_k,\beta_k)
		\begin{bmatrix}
			m_{x,k} & m_{y,k} & m_{z,k}
		\end{bmatrix}^{T}
		+
		\prescript{K}{}{\bm{m}}_{\mathrm{d},k}.
        \notag
	\end{align}
	The quasi-steady moment uses the same wind-to-wing-axis transformation as the force, whereas the damping moment is added directly in wing coordinates. The configuration-dependent geometry ratios and geometric-center location are modeled explicitly, and the identified coefficients provide additional flexibility in representing the wing hydrodynamic forces and moments.

	Let the coefficient groups associated with $K_H$ be 
	\begin{align}
		\mathcal{J}_S
		&=
		\{C_{D0},C_{D\alpha},C_{S\beta},C_{L\alpha},
		C_{m0},C_{m\alpha},C_{mq}\},\notag\\
		\mathcal{J}_{Sb}
		&=
		\{C_{x\beta},C_{z\beta}\},
		\qquad
		\mathcal{J}_{I}
		=
		\{C_{xp},C_{zr}\}.
		\label{eq:coeff_sets}
	\end{align}
	For each coefficient $C_j$, the effective hydrodynamic coefficient is scaled by coefficient-specific scaling functions
	\begin{align}
		C_{j,k}^{\mathrm{eff}}
		&=
		C_{j,k}\,K_{\mathrm{H}}(\eta_j;\beta_{j,0},\beta_{j,1}),
		\label{eq:coeff_scale}
		\\
		\eta_j
		&=
		\begin{cases}
			\eta_S, & C_j\in\mathcal{J}_S,\\
			\eta_S\eta_b, & C_j\in\mathcal{J}_{Sb},\\
			\eta_I, & C_j\in\mathcal{J}_{I}.
		\end{cases}
		\label{eq:coeff_scale_inputs}
	\end{align}
	The corresponding hydrodynamic wrench of wing $k$, referred to the hinge and expressed in the wing frame, is
	\begin{align}
		\prescript{K}{}{\bm{\tau}}_{\mathrm{h,k}}
		=
		\begin{bmatrix}
			\prescript{K}{}{\bm{f}}_{\mathrm{h,k}}\\
			\prescript{K}{}{\bm{m}}_{\mathrm{h,k}}
			+ \prescript{K}{}{\bm{r}}_{\mathrm{gc}}(\lambda_k)\times \prescript{K}{}{\bm{f}}_{\mathrm{h,k}}
		\end{bmatrix}.
		\label{eq:taukh}
	\end{align}
	The cross-product term transfers the force-induced moment from the geometric center to the hinge, explicitly incorporating the configuration-dependent moment arm.
	
	\subsection{Fuselage Hydrodynamic Model and Total External Wrench}
	The fuselage hydrodynamic damping is represented by a reduced Fossen-type model with linear and quadratic terms in the fuselage twist $\bm{\nu}_b$,
	\begin{align}
		\prescript{B}{}{\bm{\tau}}_{\mathrm{h,b}}
		=
		\bm{D}_{\nu} \bm{\nu}_b + \bm{D}_{|\nu|\nu}|\bm{\nu}_b|\bm{\nu}_b,
		\label{eq:fuselage_damping}
	\end{align}
	where the constant linear damping matrix is
	\begin{align}
		\bm{D}_{\nu}
		=
		\begin{bmatrix}
			X_u & 0 & 0 & 0 & 0 & 0\\
			0 & Y_v & 0 & 0 & 0 & Y_r\\
			0 & 0 & Z_w & 0 & Z_q & 0\\
			0 & 0 & 0 & K_p & 0 & 0\\
			0 & 0 & M_w & 0 & M_q & 0\\
			0 & N_v & 0 & 0 & 0 & N_r
		\end{bmatrix},
		\label{eq:Dl}
	\end{align}
	and $\bm{D}_{|\nu|\nu}$ shares the same sparsity pattern, with each coefficient, for example, $X_u$ replaced by its quadratic counterpart, $X_{|u|u}$. The surge and roll channels carry pure damping, whereas the sway--yaw and heave--pitch channels are cross-coupled. Hull symmetry reduces the number of independent parameters through $Z_w=Y_v$, $Z_q=-Y_r$, $N_r=M_q$, and $N_v=-M_w$. The fuselage model captures the effective viscous loads on the main body, whereas the large-range morphing-wing loads are described by the wing model above. This decomposition follows the component build-up principle \cite{fossen2011handbook}, in which body and wing contributions are modeled separately and then assembled into the total wrench.
	
	The total external wrench acting on the base dynamics is
	\begin{align}
		\prescript{B}{}{\bm{\tau}}_{\mathrm{ext}}
		=
		\prescript{B}{}{\bm{\tau}}_{\mathrm{h,b}}
		+ \prescript{B}{}{\bm{\tau}}_{\mathrm{prop}}
		+ \sum_{k\in\mathcal{W}}
		\prescript{B}{K}{\bm{X}}\prescript{K}{}{\bm{\tau}}_{\mathrm{h,k}}.
		\label{eq:tauext}
	\end{align}
	Here, $\prescript{B}{}{\bm{\tau}}_{\mathrm{h,b}}$ and $\prescript{B}{}{\bm{\tau}}_{\mathrm{prop}}$ denote the fuselage hydrodynamic and propulsion wrenches expressed in the base frame, respectively, and $\prescript{B}{K}{\bm{X}}\prescript{K}{}{\bm{\tau}}_{\mathrm{h,k}}$ maps the hydrodynamic wrench of wing $k$ to the base frame.
	
	\section{Sequential Parameter Identification}
	\label{sec:identification}

	The proposed model comprises five parameter sets, fuselage added-mass terms, fuselage hydrodynamic coefficients, wing added-mass terms, wing hydrodynamic coefficients, and configuration-dependent scaling parameters. Estimating all five sets simultaneously from raw trajectory data leads to an ill-conditioned problem, as the dynamic residual couples inertia, Coriolis, restoring, and hydrodynamic contributions. In addition, numerical differentiation of measured trajectories introduces significant noise. To address these issues, a sequential identification strategy is adopted.

	This staged identification procedure is motivated by the negligible hydrodynamic contribution of folded wings and the need for a reliable baseline model before estimating configuration-dependent effects~ \cite{wang2018three}. Accordingly, a five-stage identification procedure is adopted as outlined in Algorithm~\ref{alg:si}. Fuselage coefficients and added-mass terms are first identified from folded-wing data; wing hydrodynamic parameters are then estimated with the fuselage model fixed; finally, configuration-dependent scaling parameters are optimized with all previous parameters fixed.

	\begin{algorithm}[!t]
		\caption{Five-Stage Sequential System Identification}
		\label{alg:si}
       \renewcommand{\algorithmicrequire}{\textbf{Input:}}
       \renewcommand{\algorithmicensure}{\textbf{Output:}}
		\begin{algorithmic}[1]
			\REQUIRE Datasets $\mathcal{D}_{\mathrm{train}}$ and $\mathcal{D}_{\mathrm{test}}$, the state weights $\bm{W}$, initial physical parameters
			\ENSURE Identified fuselage coefficients, wing coefficients, added-mass terms, added-inertia terms, and scaling parameters
			\STATE Build the residual fuselage wrench $\bm{Y}_{\mathrm{fuselage}}$
			\STATE Stage A: Solve fuselage LS initialization for $\bm{\Theta}_\mathrm{bt},\bm{\Theta}_\mathrm{br}$
			\STATE Stage B: Refine fuselage hydrodynamic, added-mass, and added-inertia parameters with GD
			\STATE Build the residual wing wrench $\bm{Y}_{\mathrm{wing}}$
			\STATE Stage C: Solve wing LS initialization for $\bm{\Theta}_\mathrm{wf},\bm{\Theta}_\mathrm{wm}$
			\STATE Stage D: Refine wing hydrodynamic, added-mass, and added-inertia parameters with GD
			\STATE Stage E: Identify configuration-dependent scaling parameters $\bm{\Theta}_\mathrm{KA},\bm{\Theta}_\mathrm{KH}$ with GD
			\STATE Evaluate the identified model on $\mathcal{D}_{\mathrm{test}}$
		\end{algorithmic}
	\end{algorithm}

	\subsection{Fuselage Parameter Identification}
	\label{sec:body}
	Since the fuselage and wings have regular geometric shapes, their added-mass terms are computed analytically and used as initial values. The fuselage added mass is estimated using Lamb's ellipsoid theory~\cite{lamb1924hydrodynamics}, and the wing added mass is obtained by strip-theory integration~\cite{fossen1994guidance}. From \eqref{eq:eom}, the total external wrench $\prescript{B}{}{\bm{\tau}}_{\mathrm{ext}}$ is computed from the experimental data. For folded-wing configurations, the wing hydrodynamic contribution is negligible, i.e.,
	$\sum_{k\in\mathcal{W}}\prescript{B}{k}{\bm{X}}\bm{\tau}_{\mathrm{h,k}}=\bm{0}$.
	The residual fuselage hydrodynamic wrench is therefore written in the linear-in-parameters form
	\begin{align}
		\bm{Y}_{\mathrm{fuselage}}
		=&
		\prescript{B}{}{\bm{\tau}}_{\mathrm{ext}}
		-
		\prescript{B}{}{\bm{\tau}}_{\mathrm{prop}}
		\notag
		=
		\begin{bmatrix}
			\bm{A}_\mathrm{bf} & \bm{0} \\
			\bm{0} & \bm{A}_\mathrm{bm}
		\end{bmatrix}
		\begin{bmatrix}
			\bm{\Theta}_\mathrm{bf} \\
			\bm{\Theta}_\mathrm{bm}
		\end{bmatrix},
		\label{eq:body_ls}
	\end{align}
	where $\bm{\Theta}_\mathrm{bf}$ and $\bm{\Theta}_\mathrm{bm}$ collect the fuselage-force and fuselage-moment hydrodynamic coefficients, respectively. The regressors $\bm{A}_\mathrm{bf}$ and $\bm{A}_\mathrm{bm}$ are constructed from the measured base-frame linear and angular velocities. The initial least-squares (LS) estimate is obtained from
	\begin{align}
		\bm{\Theta}_{\mathrm{LS}}
		=
		\arg\min_{\bm{\Theta}}
		\|\bm{Y}-\bm{A}\bm{\Theta}\|_2^2.
	\end{align}
	
	In practice, the translational and rotational residuals are solved separately and then merged under fuselage symmetry constraints. The LS solution provides a parameter-scale reference but remains sensitive to acceleration noise, since $\dot{\bm{\nu}}_b$ enters the computation of $\prescript{B}{}{\bm{\tau}}_{\mathrm{ext}}$ through $\bm{M}(\bm{q}_j)$. Prior studies on AUV  identification further show that LS is computationally efficient but susceptible to measurement noise~\cite{wehbe2017experimental}. 

	To address these limitations, LS is used as a conditional warm start for gradient descent (GD) refinement. The refinement minimizes a trajectory prediction error, which evaluates state prediction errors over finite windows and avoids directly differentiating measured trajectories. Here, a trajectory refers to a complete experimental trial from the initial condition to the final state, whereas a window denotes a fixed-length temporal segment extracted from a trajectory. Accordingly, trajectory-level errors evaluate the accumulated prediction performance over an entire experiment, while window-level errors characterize local prediction accuracy over individual segments. LS coefficients are retained only when within admissible bounds; otherwise, they are re-initialized within the feasible interval. GD refines fuselage hydrodynamic, added-mass and added-inertia parameters. 
    
    For trajectory $i$, window start $j$, and sample $k$, the prediction error is defined as $\bm{e}_{i,j,k}(\bm{\Theta})=\bm{x}_{\mathrm{p}}^{i}(t_{j+k};\bm{\Theta})-\bm{x}^{i}(t_{j+k})$ where $\bm{x}_{\mathrm{p}}^{i}(t_{j+k};\bm{\Theta})$ is the model prediction initialized from the measured state at $t_j$, and $\bm{x}^{i}(t_{j+k})$ is the corresponding measurement. The estimation state is defined as $\bm{x}=[\bm{p}^T,\bm{e}^T,\bm{\nu}_b^T]^T\in\mathbb{R}^{12}$ where $\bm{p}$, $\bm{e}$, and $\bm{\nu}_b$ denote position, attitude, and fuselage twist expressed in $B$, respectively. The prediction error is defined as a normalized mean squared error (NMSE), i.e.,
	\begin{align}
		\mathcal{L}_{i,j}(L_w,\bm{\Theta})
		=&
		\frac{1}{L_w-1}
		\sum_{k=1}^{L_w-1}
		\frac{1}{n}
		\left\|
		\bm{W}\bm{e}_{i,j,k}(\bm{\Theta})
		\right\|_2^2,
		\label{eq:Jb}
	\end{align}
	where $L_w$ is the prediction window length, and $n$ is the state dimension. The weighting matrix $\bm{W}\in\mathbb{R}^{n\times n}$ performs normalization based on the min-max range of each state variable in the dataset. State propagation is implemented using a second-order Runge--Kutta integrator with $L_w=360$ samples.
	
	Training data are constructed by partitioning trajectories into non-overlapping windows, forming $\mathcal{D}$. For each  $(i,j)\in\mathcal{D}$, the model is initialized with the measured state, evaluated over the window, and updated via backpropagation of the window-level prediction error. Window shuffling is optionally applied to improve optimization robustness.	
	
	\subsection{Wing Parameter Identification}
    After identification of fuselage coefficients, the wings are deployed symmetrically to identify the wing hydrodynamic parameters, while the fuselage parameters obtained in Section~\ref{sec:body} are held fixed. A left--right symmetry assumption is adopted, consistent with prior fin-actuated robotic-fish modeling studies~\cite{wang2018three}. The wing hydrodynamic contribution is isolated by subtracting the propulsion and fuselage hydrodynamic wrenches from $\prescript{B}{}{\bm{\tau}}_{\mathrm{ext}}$, yielding a linear-in-parameter formulation, i.e.,
	\begin{align}
		\bm{Y}_{\mathrm{wing}}
		= &
		\prescript{B}{}{\bm{\tau}}_{\mathrm{ext}}
		-
		\prescript{B}{}{\bm{\tau}}_{\mathrm{prop}}
		-
		\bm{Y}_{\mathrm{fuselage}}
		\notag \\
		=&  \sum_{k\in\mathcal{W}} \begin{bmatrix}
			\bm{A}_\mathrm{wf} & \bm{0}\\
			[\bm{r}_{\mathrm{gc},k}]^{\times}\bm{A}_\mathrm{wf} &\bm{A}_\mathrm{wm}
		\end{bmatrix}
		\begin{bmatrix}
			\bm{\Theta}_\mathrm{wf} \\ \bm{\Theta}_\mathrm{wm}
		\end{bmatrix},
		\label{eq:wing_ls}
	\end{align}
	where $\bm{\Theta}_\mathrm{wf}$ and $\bm{\Theta}_\mathrm{wm}$ denote the wing force and wing moment coefficients, respectively.

	After the least-squares initialization of $\bm{\Theta}_\mathrm{wf}$ and $\bm{\Theta}_\mathrm{wm}$, the wing parameters are refined with the previously identified fuselage parameters fixed. Consistent with the fuselage refinement, the optimized parameter vector contains the wing force and moment coefficient blocks, the wing added-mass terms, and the wing added-inertia terms. The refinement minimizes the prediction error defined in \eqref{eq:Jb}.

	In the final stage, scaling parameters are identified to modulate the added mass, added inertia, and wing hydrodynamic coefficients through sweep-induced geometry ratios. Let $\bm{\Theta}_\mathrm{KA}$ and $\bm{\Theta}_\mathrm{KH}$ represent the scaling parameters associated with $K_{\mathrm{A}}(\eta;\zeta_{A})$, $K_{\mathrm{H}}(\eta;\zeta_{H,0},\zeta_{H,1})$, respectively. In this stage, all baseline fuselage and wing parameters identified in the previous stages are kept fixed, and only the scaling parameters are optimized. This staged procedure reduces the coupling among parameter sets at the beginning of training. 
	
	\section{Experiment}
	\label{sec:experiment}
	
	\subsection{Experimental Setup}
\begin{figure}[!h]
		\centering
		\includegraphics[width=\columnwidth]{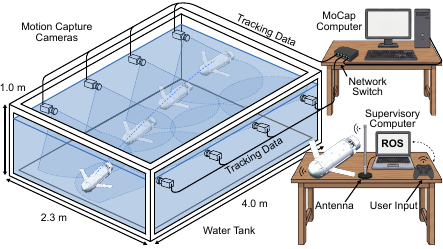}
		\caption{Experimental setup for hydrodynamic parameter identification and dynamic model validation.}
		\label{fig:experiment}
	\end{figure}

	All experiments were conducted in a $4.0\;\mathrm{m}\times2.3\;\mathrm{m}\times1.0\;\mathrm{m}$ indoor water tank equipped with eight synchronized underwater motion-capture cameras as illustrated in Fig.~\ref{fig:experiment}. The six-degree-of-freedom pose of the base frame was measured at 90~Hz using custom waterproof active LED markers mounted at known fuselage locations. The marker layout was designed to avoid symmetry ambiguity and maintain marker visibility during motion.

	The motion-capture data are used for parameter identification. Internal stepper encoders measure the actuation angles, from which the MCU computes the absorbed-water volume, translating-mass position, and rotating-mass position. These states are transmitted to the supervisory computer through the full-duplex transceiver to verify that the commanded operating conditions are achieved. 
	
 \begin{figure*}[!t]
    \centering
    \includegraphics[width=\textwidth]{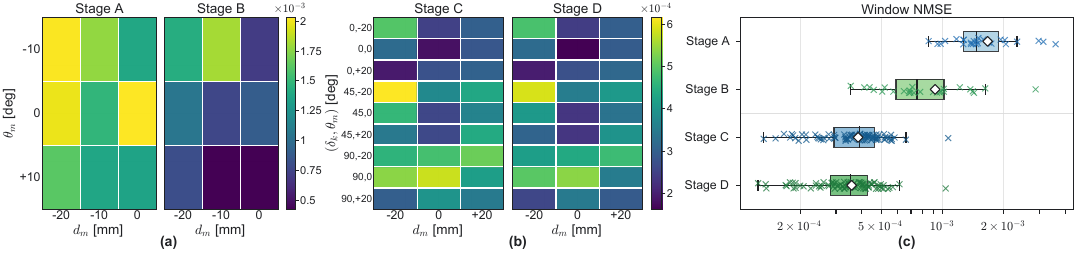}
    \caption{Model prediction errors after sequential system identification. (a) Trajectory-level mean NMSE heat maps after  fuselage-parameter identification in Stages A--B. (b) Trajectory-level mean NMSE heat maps after  wing-parameter identification in Stages C--D. (c) Window-level NMSE distributions after Stages A--D on a logarithmic scale. Crosses denote individual prediction windows, and diamonds denote the mean values.}
    \label{fig:stage_abcd_loss_composite}
\end{figure*}

	Experiments were conducted under multiple internal-ballast settings and symmetric wing-sweep angles to sufficiently excite the dynamics required by Algorithm~\ref{alg:si}. Each operating condition was repeated four times. Longitudinal ballast translation was used to excite pitch-related dynamics, while ballast rotation excited roll and lateral-force dynamics. All experiments were performed under symmetric wing-sweep conditions, i.e.,
	\begin{align}
		(\lambda_l,\lambda_r)=(-\lambda,\lambda), \quad
		\lambda \in \{0^\circ,40^\circ,80^\circ,90^\circ\},
		\label{eq:symmetric_sweeps}
	\end{align}
	where $\lambda$ denotes the commanded symmetric sweep amplitude. The case $\lambda=0^\circ$ corresponds to the fully deployed state with the maximum effective wing area, whereas $\lambda=90^\circ$ corresponds to the fully folded state with a streamlined torpedo-like profile. The intermediate sweeps exercise the scaling functions in Section~\ref{sec:hydro}. These cases are used to identify and validate the configuration-dependent parameters. Table~\ref{tab:exp_summary} summarizes the experimental settings and the number of runs.

	\begin{table}[!h]
		\centering
		\caption{Summary of experimental operating conditions}
		\label{tab:exp_summary}
		\renewcommand{\arraystretch}{1.0}
		\begin{tabular}{lccc}
			\toprule
			\textbf{Parameter} & \textbf{Condition I} & \textbf{Condition II} & \textbf{Condition III} \\
			\midrule
			Wing state & Folded & Deployed & Swept \\
			$\lambda$ [deg] & $90$ & $0$ & $40,80$ \\
			$\delta_k$ [deg] & $0$ & $0,45,90$ & $0$ \\
			$d_m$ [mm] & $(-20,-10,0)$ & $(-20,0,+20)$ & $(-20,0,+20)$ \\
			$\theta_m$ [deg] & $(-10,0,+10)$ & $(-20,0,+20)$ & $(-20,0,+20)$ \\
			Total runs & $36$ & $108$ & $72$ \\
			\bottomrule
		\end{tabular}
	\end{table}

	Measured pose and velocity were smoothed with a zero-phase fourth-order Butterworth low-pass filter. The datasets  $\mathcal{D}_{\mathrm{train}}$ and $\mathcal{D}_{\mathrm{test}}$ were split at the trajectory level, with the first three runs for training and the last run for testing under each operating condition. The folded-wing, fully deployed-wing, and multi-sweep datasets were used to identify the fuselage, wing, and scaling parameters, respectively.

\subsection{Stage A--D Model Parameter Identification}
\label{sec:stage_identification}
The prediction error was evaluated using the per-window NMSE defined in \eqref{eq:Jb}, with $L_w=360$ samples and the diagonal normalization matrix $\bm{W}$ used during training. For the stage-wise comparison, Stage A and Stage B in Algorithm~\ref{alg:si} denote fuselage LS identification and GD refinement, respectively. Stage C and Stage D denote wing LS identification and GD refinement, respectively. Stage E denotes configuration-dependent scaling identification.

The fuselage hydrodynamic, added-mass, and added-inertia parameter refinement was assessed by comparing Stage A and Stage B on 9 test trajectories containing 26 prediction windows. As shown in Figs.~\ref{fig:stage_abcd_loss_composite}(a) and (c), Stage B reduced the mean window-level NMSE from $1.666\times10^{-3}$ to $9.201\times10^{-4}$, corresponding to a 44.8\% reduction. The median NMSE decreased by 48.9\%. At the trajectory level, Stage B yielded a lower mean NMSE for eight of the nine test trajectories. This trajectory-level distribution shows that the improvement was not driven by isolated windows.

The wing hydrodynamic, added-mass, and added-inertia parameter refinement was assessed by comparing Stage C and Stage D on 27 test trajectories containing 83 prediction windows. Figs.~\ref{fig:stage_abcd_loss_composite}(b) and (c) show a smaller but consistent reduction after GD refinement. The mean window-level NMSE decreased from $3.853\times10^{-4}$ to $3.577\times10^{-4}$, corresponding to a 7.15\% reduction. The median NMSE decreased by 9.80\%. Stage D reduced the trajectory-level mean NMSE in 23 of the 27 test trajectories. These results indicate that the fuselage and wing refinements contributed complementary gains. Stage B provided the larger reduction in the overall prediction error, whereas Stage D gave a further improvement in wing-related hydrodynamic and added-mass parameters.

\subsection{Multibody Versus Single-Rigid-Body Modeling}
\label{sec:multibody_validation}
The comparison evaluated the contribution of the configuration-dependent multibody formulation. The variants were tested on the same 18 test trajectories and 41 prediction windows. The multibody variant used the identified configuration-dependent multibody dynamics and the learned Stage E scaling functions. The single-rigid-body variant is the single-body baseline with the wings fully deployed and the parameters identified in Stage B and Stage D. 

\begin{table}[!h]
\centering
\caption{Comparison of window-level and trajectory-level NMSE values for the multibody and single-rigid-body model. Values are scaled by $10^{-3}$.}
\label{tab:multibody_single_rigid}
\begin{tabular}{lcccc}
\toprule
\multirow{2}{*}{Variant}
& \multicolumn{2}{c}{Window}
& \multicolumn{2}{c}{Trajectory}  \\
\cmidrule(lr){2-3} \cmidrule(lr){4-5}
& Mean & Median & Mean & Median  \\
\midrule
Multibody           & 1.475 & 1.301 & 1.413 & 1.272\\
Single-rigid-body         & 5.738 & 5.623 & 5.715 & 5.565\\
\bottomrule
\end{tabular}
\end{table}
Table~\ref{tab:multibody_single_rigid} summarizes the prediction errors at both window and trajectory levels. Compared with the single-rigid-body baseline, the proposed multibody model reduced the window-level NMSE by 74.3\%. A consistent improvement was observed at the trajectory level, where the NMSE reduction reached 75.3\%. Furthermore, the single-rigid-body baseline exhibited larger prediction errors than the multibody model across all 41 prediction windows and 18 test trajectories, demonstrating the effectiveness of incorporating the configuration-dependent multibody dynamics. 

\subsection{Performance Analysis of Scaling Function Design}
\label{sec:scaling_ablation}
The second validation evaluated the role of individual Stage E scaling-function groups. Four ablations were tested with the same identified parameters. When a scale group was disabled, its corresponding multiplicative scaling function was set to one during evaluation. The full model retained all Bernstein scaling functions. The no-force-scale variant disabled the hydrodynamic force coefficient scales associated with $\bm{\Theta}_\mathrm{wf}$, including $C_{D0}$, $C_{D\alpha}$, $C_{S\beta}$, and $C_{L\alpha}$. The no-moment-scale variant disabled the hydrodynamic moment coefficient scales associated with $\bm{\Theta}_\mathrm{wm}$, including $C_{x\beta}$, $C_{xp}$, $C_{m0}$, $C_{m\alpha}$, $C_{mq}$, $C_{z\beta}$, and $C_{zr}$. The added-mass and added-inertia variants disabled the translational added-mass and rotational added-inertia scale groups, respectively. The ablation results are reported in Table~\ref{tab:ablation_results}.

    \begin{table}[!h]
    \centering
    \caption{NMSE comparison of the Stage E scaling-function ablation at window and trajectory levels. Values are scaled by $10^{-3}$.}
    \label{tab:ablation_results}
    \begin{tabular}{lcccc}
    \toprule
    \multirow{2}{*}{Variant}
    & \multicolumn{2}{c}{Window}
    & \multicolumn{2}{c}{Trajectory}  \\
    \cmidrule(lr){2-3} \cmidrule(lr){4-5}
    & Mean & Median & Mean & Median  \\
    \midrule
    No added-mass scale & 1.581 & 1.536 & 1.515 & 1.452\\
    No added-inertia scale & 1.505 & 1.300 & 1.461 & 1.283\\
    No force scale & 3.916 & 3.095  & 3.499 & 2.832\\
    No moment scale & 3.279 & 2.904  & 3.350 & 3.017\\
    \hline
    \end{tabular}
    \end{table}

Table~\ref{tab:ablation_results} reports the ablation results on the 18 test trajectories comprising 41 prediction windows. The full model achieved the lowest mean window-level NMSE, with a value of $1.475\times10^{-3}$. Disabling the added-inertia scale and added-mass scale increased the mean error by 2.0\% and 7.2\%, respectively. These consistent increases indicate that the added-mass and added-inertia scaling functions provide measurable corrections to the configuration-dependent inertial terms.

Removing the hydrodynamic moment or force scales caused much larger degradation, increasing the mean error by 122.3\% and 165.6\%, respectively. The force-scale ablation was worse than the full model in all 41 prediction windows and all 18 trajectory-level means, while the moment-scale ablation was worse in 38 of the 41 windows and all 18 trajectory-level means. These results indicate that the force and moment scaling functions carry the dominant hydrodynamic correction associated with wing folding, whereas the added-mass and added-inertia scales refine the prediction error. 

\subsection{Narrow-Gap Traversal With Morphing Wings}
	\begin{figure}[!t]
		\centering
		\includegraphics[width=\columnwidth]{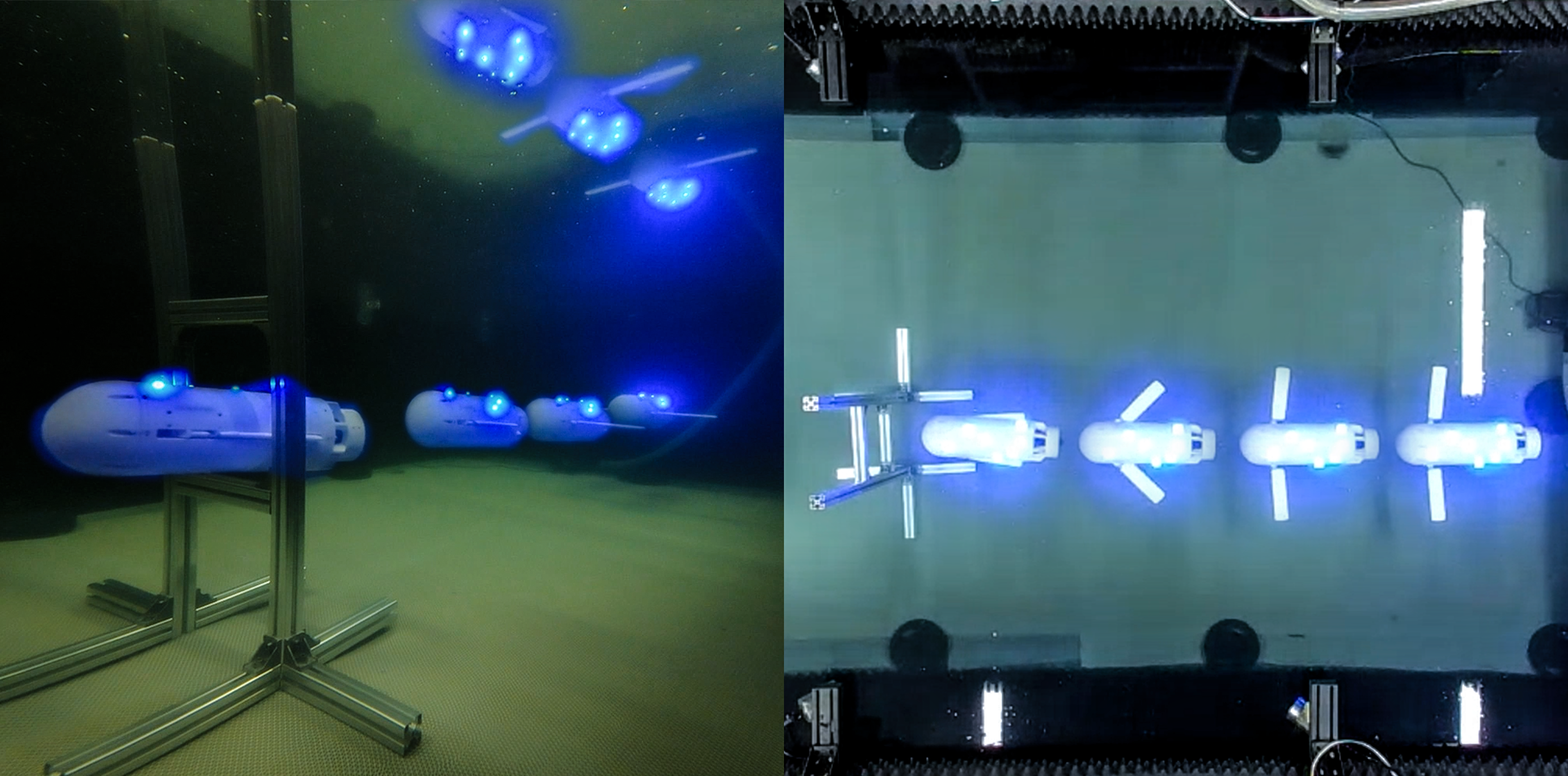}
		\caption{Experimental demonstration of morphing-wing-assisted narrow-gap traversal of FoDeGlider in a confined environment, including underwater and overhead perspectives.}
		\label{fig:openloop}
	\end{figure}
	Narrow-gap traversal experiments, as illustrated in Fig.~\ref{fig:openloop}, evaluated shape reconfiguration and maneuvering capabilities of the prototype. A 25~cm square opening was assembled from aluminum profiles with its center 40~cm above the tank bottom and 1.4~m from the short tank side. Initially, the prototype fuselage center was approximately 50~cm from the tank wall, and both wings were swept to $10^\circ$.
	A ROS node sent the predefined command sequence to the onboard controller through the wireless full-duplex transceiver. The thruster was driven with a 1.60~ms PWM pulse for 7~s, followed by a 1~s wing-folding command from $10^\circ$ to $90^\circ$. In the folded configuration, the wings retracted into the fuselage slots, reducing the maximum prototype diameter to approximately 15.5~cm, including the marker thickness. FoDeGlider completed the narrow-gap traversal task, demonstrating its unique locomotion capability in constrained aquatic environments.

	\section{Conclusion}
	This paper presented FoDeGlider, a miniature hybrid underwater glider with large-range foldable and deflectable wings and a configuration-dependent multibody dynamics model tailored to it. By parameterizing the composite inertia, velocity transport, restoring wrench, and component hydrodynamics by the wing configuration, the Fossen-form model captures the coupled effects of morphing geometry, mass redistribution, geometric-center migration, and configuration-dependent hydrodynamic parameters. A sequential identification framework with LS initialization, GD-based refinement, and scaling-parameter tuning was developed, and experiments validated the model across multiple symmetric sweep-amplitude configurations.
	The experimental results support the practical identifiability of the proposed framework within the tested sweep range. Future work will extend the experiments to asymmetric sweep configurations and examine how the identified model supports maneuver planning and closed-loop control.
	
	\bibliographystyle{IEEEtran}
	\bibliography{references}
    
    \begin{IEEEbiography}[{\includegraphics[width=1in,height=1.25in,clip,keepaspectratio]{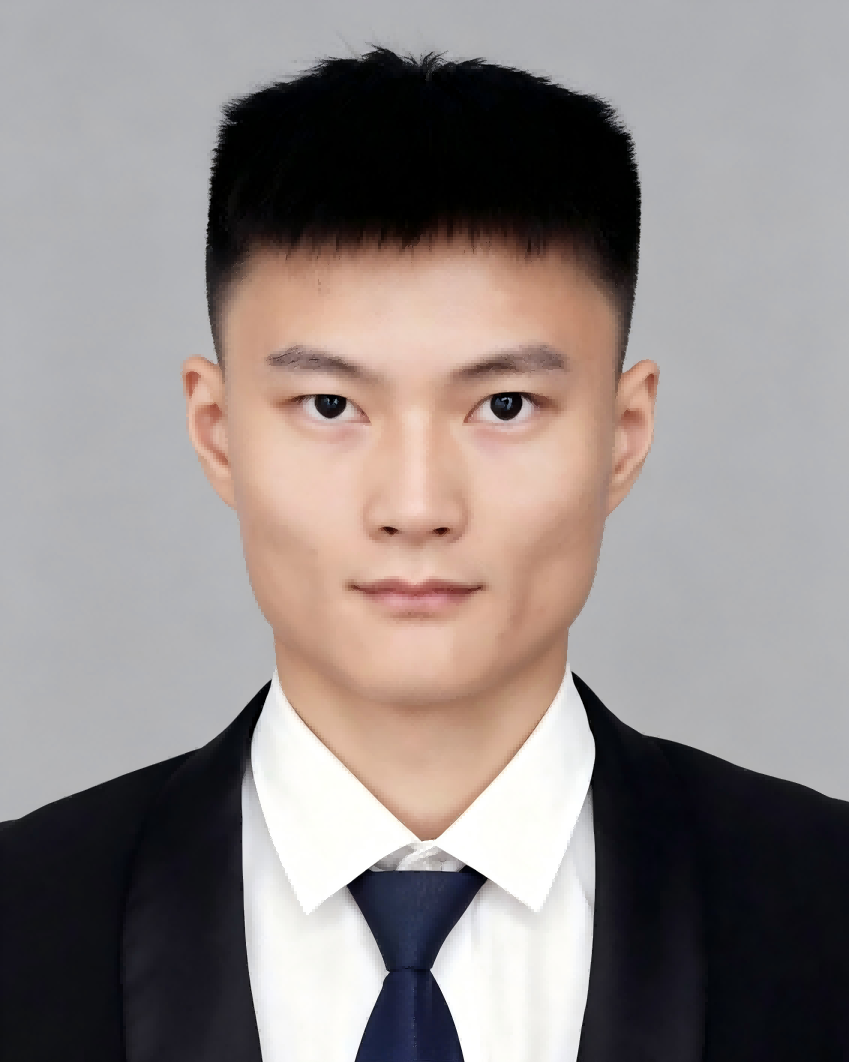}}]{Yongjian Zhu}
    received the bachelor's degree in robotics engineering in 2023 from Peking University, Beijing, China, where he is currently working toward the Ph.D. degree in general mechanics and foundation of mechanics with the School of Advanced Manufacturing and Robotics. 
    
    His current research interests include morphing gliders, dynamics modeling, and robot learning.
    \end{IEEEbiography}
    
    \begin{IEEEbiography}[{\includegraphics[width=1in,height=1.25in,clip,keepaspectratio]{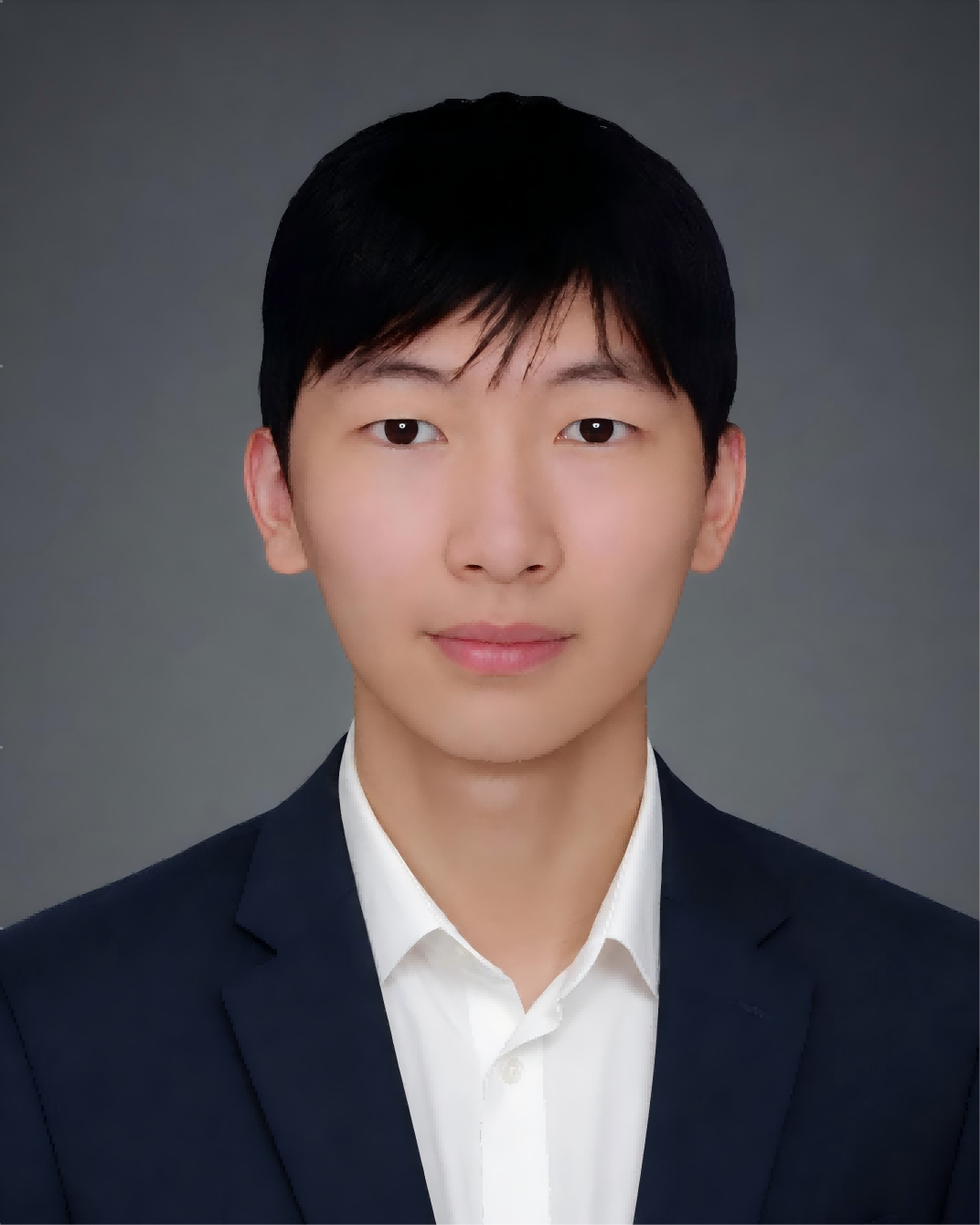}}]{Yusen Tao}
    received the bachelor's degree in robotics engineering in 2026 from Peking University, Beijing, China, where he is currently pursuing the Master of Mechanical Engineering degree with the School of Advanced Manufacturing and Robotics.
    
    His current research interests include autonomous underwater vehicles, underwater sensing and perception, and machine learning.
    \end{IEEEbiography}
    
    \begin{IEEEbiography}[{\includegraphics[width=1in,height=1.25in,clip,keepaspectratio]{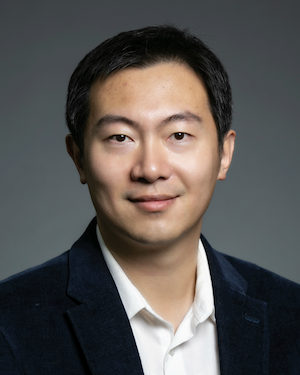}}]{Feitian Zhang}
    received the Bachelor's and Master's degrees in Automatic Control from Harbin Institute of Technology, Harbin, China, in 2007 and 2009, respectively, and the Ph.D. degree in Electrical and Computer Engineering from Michigan State University, East Lansing, MI, in 2014. 

    Feitian Zhang is currently an Associate Professor in the School of Advanced Manufacturing and Robotics at Peking University. 
    Prior to joining Peking University, he was an Assistant Professor in the Department of Electrical and Computer Engineering at George Mason University (GMU), Fairfax, VA, and the founding director of the Bioinspired Robotics and Intelligent Control Laboratory (BRICLab) from 2016 to 2021. 
    His research interests include Bioinspired Robotics, Control Systems, Artificial Intelligence, Underwater Vehicles, and Aerial Vehicles. 
    \end{IEEEbiography}
\end{document}